\pdfoutput=1

\documentclass[11pt]{article}

\usepackage[final]{acl}

\usepackage{times}
\usepackage{latexsym}
\usepackage{booktabs}
\usepackage{arydshln}
\usepackage{multirow}
\usepackage{CJKutf8}
\usepackage[T1]{fontenc}
\usepackage[utf8]{inputenc}
\usepackage{microtype}

\usepackage{inconsolata}

\usepackage{graphicx}
\usepackage{inconsolata}

\newcommand{\ctslogo}{\raisebox{3.4pt}{\includegraphics[scale=0.0105]{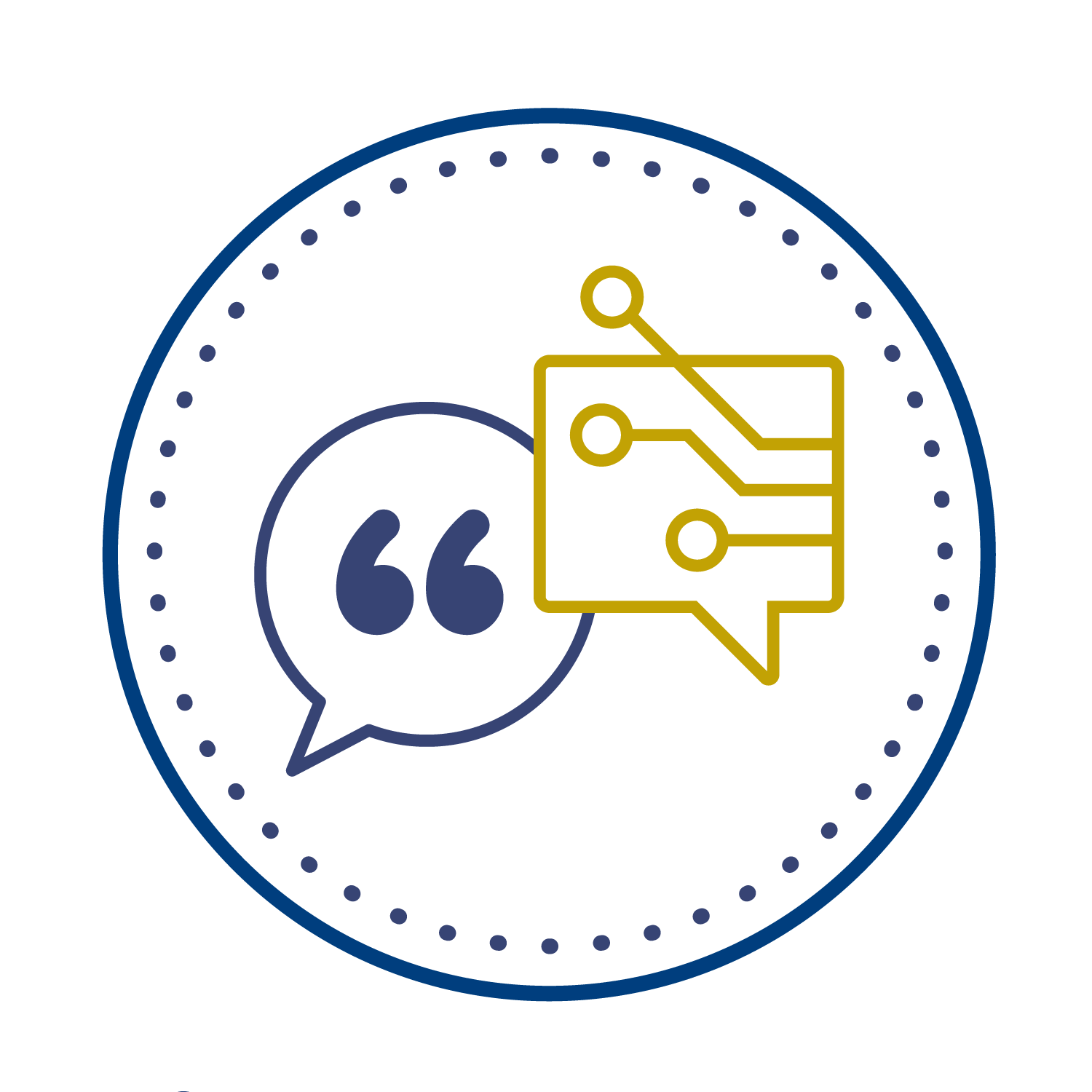}}}
\newcommand{\PAIlogo}{\raisebox{3.4pt}{\includegraphics[scale=0.080]{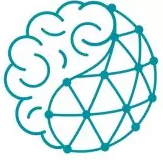}}}
\title{Automatically Generating Chinese Homophone Words to Probe Machine Translation Estimation Systems}

\author{Shenbin Qian\ctslogo, Constantin Orăsan\ctslogo, Diptesh Kanojia\PAIlogo  and Félix do Carmo\ctslogo \\
\ctslogo Centre for Translation Studies and 
\PAIlogo Institute for People-Centred AI, \\[.2em]
University of Surrey, United Kingdom \\[.13em]
\{s.qian, c.orasan, d.kanojia, f.docarmo\}@surrey.ac.uk}

\begin{document}

\maketitle
\begin{abstract}
Evaluating machine translation (MT) of user-generated content (UGC) involves unique challenges such as checking whether the nuance of emotions from the source are preserved in the target text. Recent studies have proposed emotion-related datasets, frameworks and models to automatically evaluate MT quality of Chinese UGC, without relying on reference translations. However, whether these models are robust to the challenge of preserving emotional nuances has been left largely unexplored. To address this gap, we introduce a novel method inspired by information theory which generates challenging Chinese homophone words related to emotions, by leveraging the concept of \textit{self-information}. Our approach generates homophones that were observed to cause translation errors in emotion preservation, and exposes vulnerabilities in MT systems and their evaluation methods when tackling emotional UGC. We evaluate the efficacy of our method using human evaluation for the quality of these generated homophones, and compare it with an existing one, showing that our method achieves higher correlation with human judgments. The generated Chinese homophones, along with their manual translations, are utilized to generate perturbations and to probe the robustness of existing quality evaluation models, including models trained using multi-task learning, fine-tuned variants of multilingual language models, as well as large language models (LLMs). Our results indicate that LLMs with larger size exhibit higher stability and robustness to such perturbations. We release\footnote{\url{https://github.com/surrey-nlp/homo_gen}} our data and code for reproducibility and further research. 
\end{abstract}

\section{Introduction}

\begin{figure*}[h]
  \centering
  \includegraphics[width=0.7\textwidth]{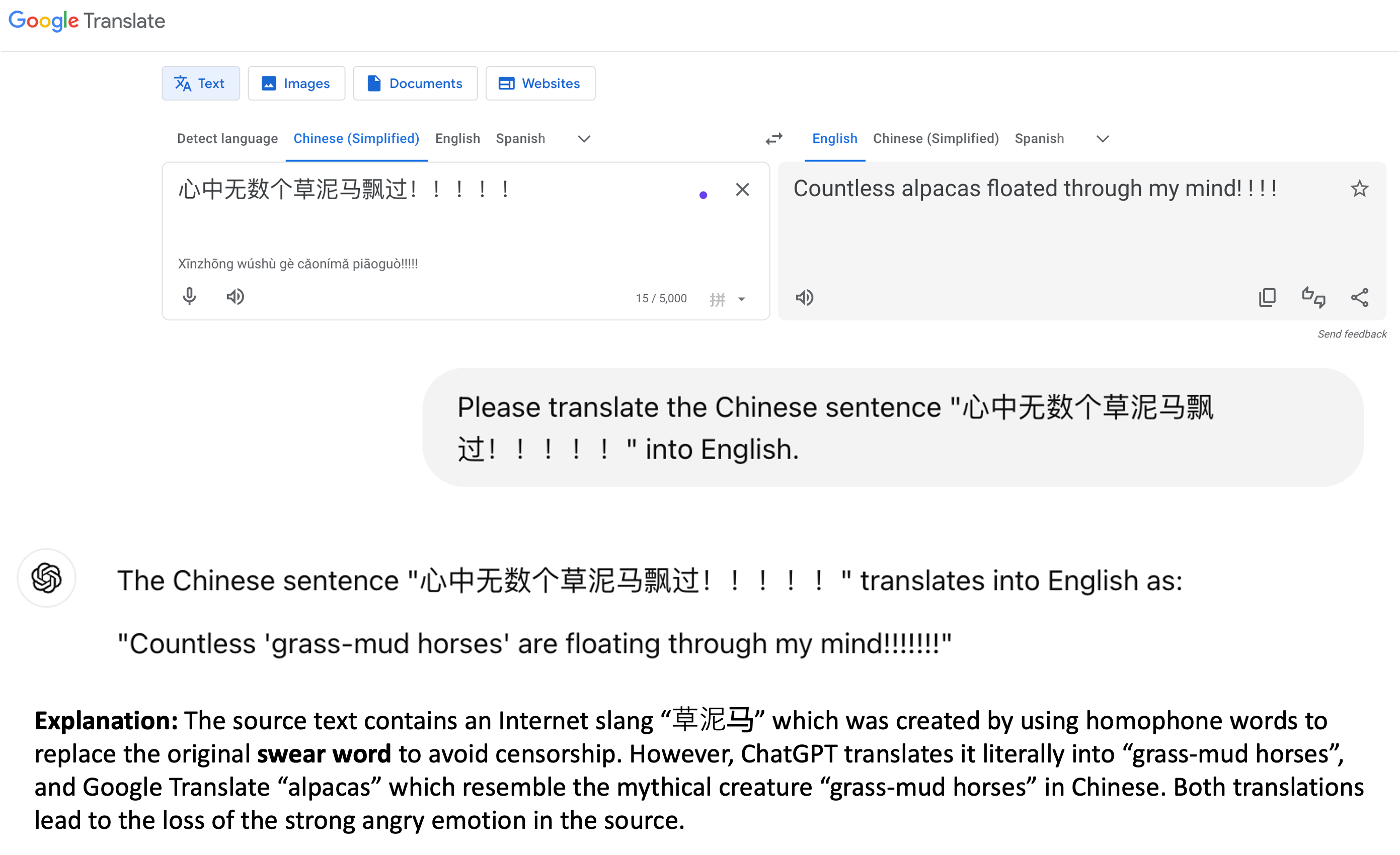}
  \caption{An example of the challenges for translating Chinese UGC}
  \label{fig.example}
\end{figure*}

Machine translation (MT) of Chinese-English news articles has been claimed to achieve human parity in recent years~\citep{Hassan2018}. However, research on machine translation of user-generated content (UGC) like tweets has revealed additional challenges including problems with handling slang, emotion, and literary devices like sarcasm and euphemisms~\citep{saadany-etal-2023-analysing}, as shown in the example translated by ChatGPT\footnote{Using \url{https://chatgpt.com/} in December 2024.} and Google Translate in Figure~\ref{fig.example}. Evaluating MT quality of such texts has become a challenging and urgent task for the improvement their translation quality~\citep{qian-etal-2024-multi}. 

Traditional ways of evaluating MT quality involve metrics such as BLEU~\citep{papineni-etal-2002-bleu}, BLEURT~\citep{sellam-etal-2020-bleurt} or BERTScore~\citep{Zhang2019} to compare the MT output with one or several reference translations. 
When references are unavailable, quality estimation (QE) methods are often used to predict scores which approximate human evaluation~\citep{Specia2018}. One approach for QE is fine-tuning multilingual pre-trained language models (PTLMs) using human evaluation scores. Frameworks like Multi-dimensional Quality Metrics (MQM)~\citep{Lommel2014}, an error-based evaluation scheme, are commonly employed to obtain the human evaluation scores for this purpose.

For machine translation of UGC, \citet{qian-etal-2023-evaluation} recruited professional translators to evaluate translations of a Chinese UGC dataset using Google Translate, based on an MQM-adapted framework. They found that homophone slang words used by netizens are the most common cause of errors in the translation of emotions. They proposed different types of QE models based on fine-tuning, multi-task learning (MTL) and large language models (LLMs) for automatic evaluation~\citep{qian-etal-2024-multi,qian-etal-2024-large-language} and claimed that their models achieved state-of-the-art performance in evaluating MT quality of UGC. In this paper, we investigate whether their models are robust enough to cope with newly generated homophones or human-improved translations. 

In this regard, we propose a method to automatically generate Chinese homophone words to probe the robustness of these QE systems towards new homophone words and human-improved translations. Our contributions can be summarized as follows: 

\begin{itemize}
    \item We leverage \textit{self-information} in information theory for the generation of Chinese homophones that can be used to replace the original word to create new slang, as a novel method. 
    \item We compare the proposed method with an existing one using \textit{percentile score}. We evaluate the two methods based on human evaluation and show that our approach achieves a higher correlation with it. 
    \item We utilize generated homophone words and human-improved translations as perturbed examples to probe existing QE models. Our analysis reveals that larger LLMs exhibit greater stability and robustness to our perturbations. 
\end{itemize}

The rest of the paper is organized as follows: Section~\ref{lit} reviews related work on quality evaluation of UGC and Chinese homophone words. Section~\ref{data} introduces the main dataset used in this study. Section~\ref{method} details the existing generation approach, our proposed method, and the human evaluation and perturbation methods. Section~\ref{results} presents and discusses the results of these evaluations. Section~\ref{conclusion} concludes the study and outlines future directions, while Section~\ref{limitations} addresses limitations and ethical considerations.


\section{Related Work} \label{lit}

Section~\ref{qe_work} provides an overview of related work on the evaluation of UGC translation, and Section~\ref{homo_gen_work} explores studies focused on Chinese UGC and the generation of homophones.

\subsection{Evaluation of UGC Translation} \label{qe_work}

Despite the tremendous improvement of translation quality since the use of neural machine translation, MT systems still struggle when translating emotion-loaded UGC such as tweets. \citet{saadany-etal-2023-analysing} analyzed machine translation of tweets for 6 language pairs and found that hashtags, slang, and non-standard orthography are the most prominent causes of translation errors. Different from the language pairs covered by \citet{saadany-etal-2023-analysing}, \citet{qian-etal-2023-evaluation} analyzed the English translation of Chinese microblog texts. They found that about 50\% of their data have translation errors in emotion preservation and about 41\% are major and critical errors. Among the causes of errors, emotion-carrying slang that contains homophones is the most frequent cause. 

To take errors in emotion into consideration during evaluation, \citet{saadany-etal-2021-sentiment} proposed a sentiment-aware measure for evaluating sentiment transfer by MT systems. Using human evaluation data based on MQM, \citet{qian-etal-2024-multi,qian-etal-2024-large-language} trained and proposed a series of QE models that can automatically assess MT quality in terms of emotion preservation. They fine-tuned and continued fine-tuned multilingual PTLMs based on TransQuest~\citep{ranasinghe-etal-2020-transquest} and COMET~\citep{rei-etal-2020-comet, stewart-etal-2020-comet, rei-etal-2022-cometkiwi}, two commonly-used QE frameworks. They also utilized the Nash \citep{pmlr-v162-navon22a} and Aligned \citep{Senushkin2023} MTL losses to train models that can perform sentence- and word-level QE concurrently. With the recent advancement of LLMs, \citet{qian-etal-2024-large-language} proposed to prompt and parameter-efficiently fine-tune LLMs for quality estimation of emotion-loaded UGC. They claimed to achieve state-of-the-art results using LLMs for evaluation. However, none of these papers answered the question: \textit{Are these models robust to new homophone slang words?} For this purpose, we propose a method to automatically generate homophone words to test the robustness of their systems. 

\subsection{Chinese Homophone Words} \label{homo_gen_work}

There have been extensive debates about what a word is in Chinese in both natural language processing and linguistic studies, as Chinese does not have a clear delimiter for word boundaries like spaces in English. Researchers have tried to define words in Chinese from different perspectives.~\citet{DiSciullo1987} defines the concept of `word' as the `listedness' characteristic of lexical items, but the `listedness' criterion fails to include many Chinese words created recently. In Chinese, usually characters, not words, are listed in lexical dictionaries. Another common way of characterizing the notion of `word' is to use semantic criteria which define a word as the smallest standalone unit that carries meaning. However, reducing concepts of a word to their semantic primitives is an extremely difficult task~\citep{Packard2000}. From a morphological perspective, a word can be defined as the output of word-formation rules in the language~\citep{DiSciullo1987}. As morphological objects are an important construct for Chinese, lots of word-like entities derived using word-formation rules but are not defined by other criteria, can be included as words by this definition. A huge amount of Internet slang created by netizens using word-formation rules such as homophone substitution can be seen as words under this definition.

Homophone substitution refers to the method which uses words or characters pronounced alike but spelt or written differently, and having different meanings from the original word or character~\citep{Meng2011}, as explained in the example \begin{CJK*}{UTF8}{gbsn}``尼玛'' \end{CJK*} in $\S$~\ref{homo_gen_method}. It is extensively used in many fields in China, such as toponymy or anthroponymy~\citep{Kałużyńska2018}, as there are so many homophones in Chinese, given it is a tonal language. Although there are studies working on this particular linguistic phenomenon~\citep{Meng2011,Chu2017,Kałużyńska2018}, to the best of our knowledge, only \citet{Hiruncharoenvate2015} have proposed a method to automatically generate homophones using percentile scores (see $\S$~\ref{homo_gen_method} for more details). In order to explore how to generate homophone words that are more likely to be used by netizens, we propose to use \textit{self-information}~\citep{Shannon1948} based on the log probability from language models. We compare our method with the existing one via human evaluation, and utilize those generated homophones as perturbations to test the robustness of QE systems proposed by \citet{qian-etal-2024-multi,qian-etal-2024-large-language}. 

\section{Data} \label{data}

We used the Human Annotated Dataset for Quality Assessment of Emotion Translation (HADQAET)\footnote{\url{https://github.com/surrey-nlp/HADQAET}} from~\citet{qian-etal-2023-evaluation} to sample UGC that contains Chinese homophone slang for automatic generation. HADQAET was chosen because, 1) its source texts contain many homophone slang; 2) it has quality evaluation data such as QE scores for the MT texts, error words related to emotion preservation and reference translations, and 3) there are QE systems trained on it (explained in $\S$~\ref{QE_models}). 

The source texts of HADQAET originated from the dataset released by the~\textit{Evaluation of Weibo Emotion Classification Technology on the Ninth China National Conference on Social Media Processing} (SMP2020-EWECT). It originally has a size of 34,768 instances. Each instance is a tweet-like text segment in Chinese, which was manually annotated with one of the six emotion labels, \textit{i.e.},~\textit{anger},~\textit{joy},~\textit{sadness},~\textit{surprise},~\textit{fear} and~\textit{neutral}~\citep{Guo2021}. \citet{qian-etal-2023-evaluation} randomly kept 5,538 instances and used Google Translate to translate them to English. To evaluate translation quality for emotion preservation, they proposed an emotion-related MQM framework and recruited two professional translators to annotate errors and their corresponding severity. Words/characters in both source and target that cause errors were highlighted for error analysis. In addition, they hired a translation company to post-edit the MT output to get reference translations~\citep{qian-etal-2024-evaluating}. More details about HADQAET can be found in~\citet{qian-etal-2023-evaluation}. 

We tokenized the source texts using~\textit{jieba}~\citep{jieba2013} and extracted the words that were highlighted as causes of error. Following~\citet{qian-etal-2023-evaluation}, we made a frequency list of these error words and picked those that contain homophone slang with a frequency higher than 10, under the supervision of a Chinese native speaker. This produced a list of 5 homophone slang words (as shown in Table~\ref{tab:homophone_errors}) that are most likely to cause translation errors. They were used in this paper to generate homophones that can be used interchangeably in the original source text. We selected the instances (167 in total) containing the 5 homophone slang words from HADQAET, including the source, MT outputs, evaluation data and reference translations to probe trained QE systems and test how robust they are. Methods for homophone generation are presented in $\S$~\ref{homo_gen_method}. Methods to create perturbed data for robustness test are described in $\S$~\ref{perturb_method}.

\begin{CJK*}{UTF8}{gbsn}

\begin{table}[]
\centering
\resizebox{8cm}{!}{%
\begin{tabular}{ccc}
\toprule
Homophone Slang Causing Errors & Human Translation & Frequencies \\ \hline
尼玛 (nima) & (f**k) your mother & 60 \\
特么 (tama) & what's the f**k & 51 \\
卧槽 (wocao) & f**k & 22 \\
草泥马 (caonima) & f**k your mother & 22 \\
劳资 (laozi) & I & 12 \\
In total & / & 167 \\
\bottomrule
\end{tabular}%
}
\caption{Homophone slang words that cause translation errors and their frequencies in HADQAET.}
\label{tab:homophone_errors}
\end{table}

\end{CJK*}

\section{Methodology} \label{method}

This section presents our methodology for homophone generation and the evaluation of generated homophones in $\S$~\ref{homo_gen_method} and $\S$~\ref{homo_eval_method}, respectively. QE models for robustness test as well as the perturbation methods are elaborated in $\S$~\ref{QE_models} and $\S$~\ref{perturb_method}.

\subsection{Homophone Generation} \label{homo_gen_method}

\begin{table}[h]
\centering
\resizebox{8cm}{!}{%
\begin{tabular}{l}
\toprule
Algorithm 1 Homophone generation \\ \hline
\textbf{Input}: $W:$ words for which to generate homophone \\
\textbf{Output}: $\tilde W:$ homophones of $W$ \\
\textbf{Candidate}: $C:$ a set of character combinations \\ that might be $\tilde W$,~\textit{i.e.} $\tilde W \in C$ \\
\textbf{Corpus}: $D:$ dictionary of character frequency in Weibo \\
\textbf{For} $w_i$ in $W$ \textbf{do} \\
\ \ \ \ ${w_i}_{root} \leftarrow Latinize(w_i)$\\
\ \ \ \ $C_{w_i} = \{Concat(DeLatinize(c_{w_i}^j)\ for\ c_{w_i}^j\ in\ {w_i}_{root})\}$\\
\ \ \ \ Optional: $C_{w_i} \leftarrow$ filter $C_{w_i}$ by $D$ \\
\ \ \ \ $\tilde W \leftarrow pick(C_{w_i})$ \\
\textbf{End for} \\
\textbf{Return} $\tilde W$ \\
\bottomrule
\end{tabular}%
}
\label{tab:algo_homo1}
\end{table}

\begin{CJK*}{UTF8}{gbsn}
The method to generate homophone words is shown in Algorithm 1. Since Chinese is a logographic language, we need to Latinize Chinese words into alphabets to get their pronunciation. For example, we can convert the slang ``尼玛'' (see Table~\ref{tab:homophone_errors} for its meaning) into ``nima'' using~\textit{Pinyin}, a system to transcribe Mandarin Chinese sounds into Latin alphabets. The Latinized words such as ``nima'', which are the root sounds/words (denoted as ${w_i}_{root}$) of the original words, can correspond to many different Chinese written words\footnote{The root sound has four different tones. Each corresponds to many different characters/words.}. We can easily generate numerous different character combinations that bear the same or similar sounds (with different tones) using the root sounds. However, many of them may not make sense and are unlikely to be used in real-world scenarios. We call them candidates (denoted as $C_{w_i}$) of our final output. We introduced a $pick()$ function explained in the following subsections to select those that are more likely to be used by netizens.
\end{CJK*}

\paragraph{Generation of Candidates}

\begin{CJK*}{UTF8}{gbsn}
After Latinization, we get the root sound of each Chinese character in the original word,~\textit{i.e.}, $c_{w_i}^j$. We gathered all Chinese characters (logographs) of the same root sound (Latin alphabets) by using the Chinese character dictionary in~\textit{jieba} for de-Latinization. A simple concatenation of each character in the same word can lead to a set of candidates, $C_{w_i}$. For example, the slang word ``尼玛'' has two characters, ``尼'' \textit{ni} and ``玛'' \textit{ma}, and each has a long list of homophone characters such as ``你'' or ``泥'' for \textit{ni} and ``吗'' or ``嘛'' for \textit{ma}. To reduce the number of candidates, we first created a dictionary (denoted as $D$) of character frequency using the full SMP2020-EWECT corpus. Then we selected character combinations whose frequency are higher than 100 to filter out those infrequent words. This resulted in a set of 172 candidates (34.4 for each) of the 5 selected homophone slang that frequently cause translation errors in emotion preservation.
\end{CJK*}

\paragraph{Picking Candidates by Percentile Score}

We used the method proposed by \citet{Hiruncharoenvate2015} as our baseline to pick candidates, which is explained in Algorithm 2. For each candidate $h$ in the set $C_{w_i}$, we summed up the frequency of each character $c_h^i$ in candidate/hypothesis $h$, using the frequency dictionary $D$. We ranked them by the aggregate frequency $F_h$ in an ascending order for each of the 5 selected slang words. The percentile score $P_{score^{w_i}}$ can be computed by dividing the index of a candidate in $C_{{w_i}^{sorted}}$ by the number of candidates in it and multiplying 100. The output homophone words can be generated by picking the top $k$ samples. 

\begin{table}[h]
\centering
\resizebox{8cm}{!}{%
\begin{tabular}{l}
\toprule
Algorithm 2 Picking candidates by percentile score \\ \hline
\textbf{Input}: $C:$ sets of candidates for $w_i$ in $W$\\
\textbf{Output}: $\tilde W:$ generated homophones \\
\textbf{Corpus}: $D:$ dictionary of character frequency in Weibo \\
\textbf{For} $h$ in $C_{w_i}$ \textbf{do} \\
\ \ \ \ $F_h = \sum_{i=1}^N freq(c_h^i)$ for $c_h^i$ in $h$, where $c_h^i \in D$ \\ 
\textbf{End for} \\
$C_{{w_i}^{sorted}} \leftarrow$ sort $C_{w_i}$ by $F_h$\\
$P_{score^{w_i}} = \{\frac{index}{length(C_{{w_i}^{sorted}})}*100\ for\ index\ in\ C_{{w_i}^{sorted}}\}$\\
$\tilde W \leftarrow P_{score^{w_i}}[1:k]$ \\

\textbf{Return} $\tilde W$ \\
\bottomrule
\end{tabular}%
}
\label{tab:algo_homo2}
\end{table}

\paragraph{Picking Candidates by Self-information}

We propose to pick candidates by self-information as shown in Equation~\ref{eq:self-info}, where $P(x)$ is the probability of an event $x$ (a word in the candidates in our case) and $I(x)$ is the self-information, which quantifies how informative an event is. Our assumption is that the generated word should be informative and unique, and at the same time not infrequent. We employed language models including the Chinese RoBERTa~\citep{cui-etal-2020-revisiting} and the Qwen1.5 series (1.8B, 4B and 7B) models~\citep{Qwen2024} to get the log probability for our candidates. 

\begin{equation} \label{eq:self-info}
    I (x) = - \log_2(P(x))
\end{equation}

\subsection{Evaluation of Homophone Words} \label{homo_eval_method}

We recruited two annotators who are frequent users of the Chinese microblogging platform, \textit{Weibo} to rate the 172 generated homophone words from 1 to 5. A score of 5 means the generated homophone can completely replace the one in the original text. A score of 1 means it can not replace the original one at all. A score of 3 is somewhere in between, meaning that the generated homophone can replace the original one, but it may take time for some readers to accept such usage.

The human evaluation was carried out in two scenarios: with (given the source microblog text) and without context (given the generated homophone along with its original word) to test if context is a factor that influences the effectiveness of the generated homophones. 

We used the Spearman correlation score~\citep{Spearman1904} to measure how the percentile and the self-information scores are correlated with the human rated scores to compare between the two methods. We also computed the Spearman correlation score between the scores of the two human annotators for references (see $\S$~\ref{eval_homo-results} for results).

To provide a quantitative complement to human evaluation, we fine-tuned the Chinese RoBERTa$_{large}$ model~\citep{cui-etal-2020-revisiting} on the SMP2020-EWECT dataset, creating an emotion classifier that achieved a macro F1 score of 0.95. Manual validation of 100 random samples confirmed the classifier's reliability, yielding an F1 score of 0.90. We then used this classifier to assess whether the predicted emotion labels remained consistent when original homophone slang was replaced with our generated homophone words. 

\subsection{QE Models for Robustness Test} \label{QE_models}

Since models proposed by~\citet{qian-etal-2024-multi,qian-etal-2024-large-language} were all trained on HADQAET, we selected two fine-tuned (FT) models based on TransQuest and COMETKIWI~\citep{rei-etal-2022-cometkiwi} respectively, one continued fine-tuned (CFT) model based on TransQuest, two MTL models based on the Nash loss, and two instruction-tuned LLMs including Mixtral-8x7B~\citep{Jiang2024} and Deepseek-67B\footnote{\url{https://www.deepseek.com/}}, as well as two parameter-efficiently fine-tuned LLMs using QLoRA~\citep{Dettmers2023-dz},~\textit{i.e.}, FT-Yi-34B and FT-Deepseek-67B. They were selected to test how robust QE models are in terms of the newly generated homophone slang words. 

\subsection{Perturbation Methods} \label{perturb_method}

We propose two perturbation methods to test the robustness of the selected QE models.

\subsubsection{Method 1: Robustness to Homophones} \label{method1}

Method 1 is to test the robustness of the QE models to our generated homophones, which were among the most frequent causes of translation errors. 

We selected the 167 instances from HADQAET that contain the 5 slang words in the source and replaced them with top 5 generated homophone words in human evaluation (see Table~\ref{tab:generated-homo} in Appendix~\ref{sec:appendix}). Everything else remained unchanged. This led to 5 groups of the 167 instances, namely, \textbf{M1G1} to \textbf{M1G5}\footnote{G1 to G5 are in a ranked order based on human evaluation.}. The QE scores produced by the selected models for the 5 groups should be more or less the same as the scores of the original source-MT group, namely, \textbf{G0}, if the models are robust. We compared the Spearman and Pearson's correlation scores among the groups for evaluation. 

\subsubsection{Method 2: Robustness to Improved Translations} \label{method2}

Method 2 is to test the robustness of the QE models to translations of improved quality.  

We asked a professional translator to correct only the translation of the homophone slang in the MT output for these 167 instances to form a perturbation group, \textit{i.e.}, \textbf{M2G1}. We also replaced the entire MT output with a human reference translation for the selected instances to form another perturbation group, \textit{i.e.}, \textbf{M2G2}. M2G1 and M2G2 are used to compare with \textbf{G0} to see the increase of QE scores, since theoretically better translations should have higher QE scores.

We calculated the percentage of the instances that see an increase of QE scores produced by the selected models to evaluate their robustness to translations of improved quality. 

\section{Results and Discussion} \label{results}

This section presents and discusses the results of evaluation of our generated homophone words as well as the results of our perturbation methods.

\subsection{Evaluation of Generated Homophones} \label{eval_homo-results}

We conducted human evaluation of the generated homophone words under two scenarios: \textbf{with} and \textbf{without} context. Results are displayed in Tables~\ref{tab:with_context} and~\ref{tab:without_context}, respectively.

\begin{table}[h]
\centering
\resizebox{7.5cm}{!}{%
\begin{tabular}{cccc}
\toprule
Methods & Annotator 1 & Annotator 2 & Avg \\ \hline
\begin{tabular}[c]{@{}c@{}} $I$ using Chinese RoBERTa\end{tabular} & 0.1257 & 0.1205 & 0.1304 \\
\begin{tabular}[c]{@{}c@{}} $I$ using Qwen1.5-1.8B\end{tabular} & 0.1957 & 0.1938 & 0.1952 \\
\begin{tabular}[c]{@{}c@{}} $I$ using Qwen1.5-4B\end{tabular} & 0.2251 & 0.2040 & 0.2215 \\
\begin{tabular}[c]{@{}c@{}} $I$ using Qwen1.5-7B\end{tabular} & \textbf{0.2799} & \textbf{0.2300} & \textbf{0.2647} \\
Percentile score & -0.0220 & -0.1219 & -0.0877 \\
\bottomrule
\end{tabular}%
}
\caption{Spearman correlation scores of self-formation ($I$) obtained on the Chinese RoBERTa, Qwen1.5 series models and the percentile score with scores annotated \textbf{with} context by Annotator 1, 2 and their average.}
\label{tab:with_context}
\end{table}

\paragraph{With Context}

We can see from Table~\ref{tab:with_context} that the Spearman correlation scores of the percentile score method are extremely low for scores of both annotators and the average score. Our self-information method improves the correlation with human annotators remarkably. This is particularly obvious when we used larger models to get the log probability, since Spearman correlation scores increase steadily when larger models are used.  

We also computed the Spearman correlation score between the two annotators as a reference to human-level correlation. Spearman correlation for the human rated scores is $0.6441$, which is still higher than our method using self-information.

\begin{table}[h]
\centering
\resizebox{7.5cm}{!}{%
\begin{tabular}{cccc}
\toprule
Methods & Annotator 1 & Annotator 2 & Avg \\ \hline
\begin{tabular}[c]{@{}c@{}} $I$ using Chinese RoBERTa\end{tabular} & 0.2050 & 0.3160 & 0.3018 \\
\begin{tabular}[c]{@{}c@{}} $I$ using Qwen1.5-1.8B\end{tabular} & 0.1837 & 0.3475 & 0.2867 \\
\begin{tabular}[c]{@{}c@{}} $I$ using Qwen1.5-4B\end{tabular} & 0.2197 & 0.3550 & 0.3156 \\
\begin{tabular}[c]{@{}c@{}} $I$ using Qwen1.5-7B\end{tabular} & \textbf{0.2379} & \textbf{0.3743} & \textbf{0.3286} \\
Percentile score & 0.0867 & 0.1516 & 0.1537 \\
\bottomrule
\end{tabular}%
}
\caption{Spearman correlation scores of self-formation ($I$) obtained on the Chinese RoBERTa, Qwen1.5 series models and the percentile scores with scores annotated \textbf{without} context by Annotator 1, 2 and their average.}
\label{tab:without_context}
\end{table}

\begin{table}[h]
\centering
\resizebox{7.5cm}{!}{%
\begin{tabular}{ccccc}
\toprule
\textbf{Group} & \textbf{Precision} & \textbf{Recall} & \textbf{F1 Score} & \textbf{Same Label} \\ \hline
M1G1 & 0.8892 & 0.8862 & 0.8675 & 0.8863 \\
M1G2 & 0.8976 & 0.9042 & 0.8904 & 0.9042 \\
M1G3 & 0.8618 & 0.8802 & 0.8634 & 0.8802 \\
M1G4 & 0.8192 & 0.8802 & 0.8480 & 0.8802 \\
M1G5 & 0.8860 & 0.8862 & 0.8764 & 0.8862 \\
\bottomrule
\end{tabular}%
}
\caption{Precision, recall, F1 score and percentage of instances that have the same label (same label) compared with the original human-annotated emotion label.}
\label{tab:emotion_label}
\end{table}

\begin{table*}[h]
\centering
\resizebox{15cm}{!}{%
\begin{tabular}{ccccccccccc}
\toprule
\multirow{2}{*}{Groups} & \multicolumn{2}{c}{FT-COMETKIWI} & \multicolumn{2}{c}{FT-TransQuest} & \multicolumn{2}{c}{CFT-TransQuest} & \multicolumn{2}{c}{MTL-XLM-V$_{base}$} & \multicolumn{2}{c}{MTL-XLM-R$_{large}$} \\
 & $\rho$ & $r$ & $\rho$ & $r$ & $\rho$ & $r$ & $\rho$ & $r$ & $\rho$ & $r$ \\ \hline
G0 & 0.2617 & 0.3211 & 0.2518 & 0.2954 & 0.2853 & 0.3219 & 0.2179 & 0.2139 & 0.1958 & 0.1841 \\
M1G1 & -3.59\% & -6.13\% & +5.79\% & +2.51\% & -8.55\% & -7.21\% & -1.83\% & +5.03\% & -2.30\% & -95.88\% \\
M1G2 & -0.50\% & -4.05\% & +8.21\% & +6.43\% & -5.06\% & -4.90\% & +13.58\% & +10.26\% & -2.76\% & -22.98\% \\
M1G3 & +2.94\% & +1.99\% & +0.77\% & +2.20\% & -9.46\% & -6.40\% & +1.29\% & -3.23\% & -5.61\% & +1.30\% \\
M1G4 & +5.85\% & +3.21\% & +7.17\% & +9.62\% & -16.57\% & -13.37\% & +11.92\% & +6.79\% & -2.20\% & +1.09\% \\
M1G5 & +1.34\% & -3.45\% & +9.99\% & +7.74\% & -14.09\% & -12.84\% & +0.09\% & +4.67\% & +0.82\% & -49.17\% \\
\bottomrule
\end{tabular}%
}
\caption{Spearman $\rho$ and Pearson's $r$ correlation scores of the perturbation groups in Method 1 on fine-tuned COMETKIWI (FT-COMETKIWI), TransQuest (FT-TransQuest) and continued fine-tuned TransQuest (CFT-TransQuest) models, and MTL models based on XLM-V$_{base}$ and XLM-R$_{large}$. The values for M1G1--M1G5 are percentage changes compared to G0. Original values can be found in Table~\ref{tab:perturb1_ft_orig} in Appendix~\ref{sec:appendix}.}
\label{tab:perturb1_ft}
\end{table*}

\begin{table*}[h]
\centering
\resizebox{13cm}{!}{%
\begin{tabular}{ccccccccc}
\toprule
\multirow{2}{*}{Groups} & \multicolumn{2}{c}{Mixtral 8x7B} & \multicolumn{2}{c}{Deepseek-67B} & \multicolumn{2}{c}{FT-Yi-34B} & \multicolumn{2}{c}{FT-Deepseek-67B} \\
 & $\rho$ & $r$ & $\rho$ & $r$ & $\rho$ & $r$ & $\rho$ & $r$ \\ \hline
G0   & 0.1886 & 0.1984 & 0.2073 & 0.1338 & 0.3413 & 0.3485 & 0.2802 & 0.2469 \\
M1G1 & -51.73\% & -131.45\% & -8.39\% & -34.39\% & +21.09\% & +21.18\% & -17.20\% & +11.62\% \\
M1G2 & -59.97\% & -131.45\% & -26.04\% & -54.52\% & -23.22\% & -22.85\% & -4.43\%  & -1.01\% \\
M1G3 & -71.46\% & -58.79\%  & -4.63\%  & +7.70\%  & -14.47\% & -12.51\% & -6.53\%  & +12.60\% \\
M1G4 & -60.18\% & -84.36\%  & -56.35\% & -96.49\% & -16.86\% & -18.94\% & +25.91\% & +55.24\% \\
M1G5 & -35.41\% & -131.35\% & -37.83\% & +0.30\%  & -44.92\% & -36.41\% & +5.71\%  & +33.86\% \\
\bottomrule
\end{tabular}%
}
\caption{Spearman $\rho$ and Pearson's $r$ correlation scores of the perturbation groups in Method 1 on LLMs and fine-tuned (FT) LLMs as listed in Section~\ref{QE_models}. For M1G1--M1G5, values are expressed as percentage changes relative to G0. Original values can be found in Table~\ref{tab:perturb1_llms_orig} in Appendix~\ref{sec:appendix}.}
\label{tab:perturb1_llms}
\end{table*}

\paragraph{Without Context}

Table~\ref{tab:without_context} re-affirms our results in Table~\ref{tab:with_context}: the self-information method obvious surpasses the percentile score method in Spearman correlation for all language models.

The Spearman score for the human rated scores without context is $0.6367$, which is similar to that of with context, but is closer to our self-information method ($0.3286$), compared with the evaluation with context ($0.6441$~\textit{vs} $0.2647$). This may be because Chinese is a context-dependent language~\citep{Stallings1975} and adding context to the generated homophone words might have an impact on the understanding of their individual meaning. 

\paragraph{Emotion Label after Replacement}
We predicted the emotion label of the 167 instances that have been replaced with the 5 generated homophone words in M1G1 to M1G5 in $\S$~\ref{method1}. Results are shown in Table~\ref{tab:emotion_label}.

Table~\ref{tab:emotion_label} indicates that the F1 scores of all groups are very close to the human validated score ($0.90$) of the emotion classifier. Close to $90$\% of the instances remain the same emotion label as that of the original source text before homophone replacement. This indicates that our generated homophone words evaluated by human annotators are reliable in terms of predicting the emotion labels.

\subsection{Results of Perturbation Methods}

\paragraph{Method 1}

Tables~\ref{tab:perturb1_ft} and~\ref{tab:perturb1_llms} show the results of our perturbation Method 1,~\textit{i.e.}, whether QE models trained by \citet{qian-etal-2024-multi,qian-etal-2024-large-language} are robust or stable to the generated homophone words, which are most frequent in causing translation errors. 

Table~\ref{tab:perturb1_ft} presents results obtained on fine-tuned (FT) COMETKIWI, fine-tuned (FT) TransQuest and continued fine-tuned (CFT) TransQuest models as well as MTL models based on XLM-V$_{base}$~\citep{Liang2023-hb} and XLM-R$_{large}$~\citep{conneau-etal-2020-unsupervised}. In each model, \textbf{G0} serves as a baseline or reference, but we also assess how stable the scores remain across \textbf{M1G1} to \textbf{M1G5} by reporting how much the score has changed in relation to G0 in percentages. For instance, if an M1G1 correlation score deviates greatly from G0 or from the adjacent group M1G2, we consider that ``fluctuation''. We can see from the table that Spearman correlation scores of M1G1-M1G5 for MTL models, especially MTL-XLM-R$_{large}$, fluctuate less than those of the FT or CFT models. This indicates that they are relatively more stable in predicting QE scores when tested with the generated homophone words.

Table~\ref{tab:perturb1_llms} shows results obtained on LLMs, including prompting LLMs for quality evaluation and fine-tuning (FT) LLMs as quality evaluators. We observe that using LLMs for QE is less stable in terms of score prediction. When we replace the original slang with our generated ones in the source, the correlation scores tend to fluctuate more than those of fine-tuned or MTL models.  Among these LLMs, larger models seem to be better at generating consistent QE scores than smaller ones, since Spearman scores of Deekseek-67B or its fine-tuned version fluctuate less than those of Mixtral 8x7B and FT-Yi-34B among the perturbation groups. 

\begin{table}[h]
\centering
\resizebox{7cm}{!}{%
\begin{tabular}{ccc}
\toprule
Models & M2G1 (\%) & M2G2 (\%) \\ \midrule
FT-COMETKIWI & 23.35 & 53.29 \\
FT-TransQuest & 45.86 & 56.35 \\
CFT-TransQuest & 33.15 & 45.30 \\
MTL-XLM-V$_{base}$ & 49.72 & 35.91 \\
MTL-XLM-R$_{large}$ & 75.69 & 67.40 \\
Mixtral   8x7B & 67.40 & 63.54 \\
Deepseek-67B & 56.91 & 74.03 \\
FT-Yi-34B & 85.64 & 83.98 \\
FT-Deepseek-67B & 81.77 & 89.50 \\
\bottomrule
\end{tabular}%
}
\caption{Percentage of instances that see a QE score increase after the MT output was improved as described in Method 2.}
\label{tab:method2}
\end{table}

\paragraph{Method 2}

Table~\ref{tab:method2} displays the percentage of instances that see an increase of the predicted QE scores after replacing the MT output with improved translations. 

Since MT outputs in M2G2 were replaced with reference translations, the percentage of instances that have increased predicted scores should be higher than those of M2G1, where only translation of the homophone slang was corrected. Comparing between the two groups, we find that for fine-tuned COMETKIWI and TransQuest models, though the percentages are usually lower than 50\%, they are higher in M2G2 than in M2G1. Whereas for MTL models, the percentages of instances that have increased scores in M2G2 are lower than those of M2G1, indicating that they are less robust towards improved translations. For LLMs, larger models such as Deepseek-67B and its fine-tuned version see an increase of the percentage of the instances that have increased scores for M2G2, whereas smaller models do not. 

Among all these QE models, LLMs such as FT-Yi-34B and FT-Deepseek-67B are more likely to produce increased scores when the translation quality is improved, like the cases in M2G1 and M2G2, since more than half of the instances experienced a score increase. This is consistent with the results from Table~\ref{tab:perturb1_llms}, which suggest that LLMs are prone to change their score prediction when the input has been changed. LLMs with large size outperform other QE models in two ways: they better reflect improvements in machine translation quality, and they maintain consistent scores when original homophone slang in the source text is replaced with generated alternatives.

\subsection{Discussion}

We observe that although our LLM-based self-information method lags behind human evaluation, it is much better than the existing percentile score method for automatically generating Chinese homophone words. Due to the context-dependent nature of the Chinese language, correlation scores to human evaluation with context can be lower than those of without context. More experiments and examples are needed for the validation of this point. 

When assessing the robustness of QE models, we find that LLM-based QE models are more likely to change their prediction scores when the input is changed. When the translation quality is improved, they are more likely to produce increased scores than fine-tuned COMETKIWI or TransQuest models or MTL models. However, when the original homophone words are replaced with our generated ones (for which human evaluation indicates they are acceptable), LLM-based models are more likely to change their predicted scores as well. LLMs with a larger size such as DeepSeek-67B and its fine-tuned versions achieved a good balance between producing consistent scores to generated homophone words and increased scores to improved translations, exhibiting great stability and robustness to our perturbations in all groups. 

\section{Conclusion and Future Work} \label{conclusion}

This paper investigates how robust emotion-related QE systems are towards emotion-loaded homophone words. For this purpose, we proposed to use self-information to automatically generate and select Chinese homophone words that frequently cause translation errors. We evaluated the efficacy of our method based on human evaluation and compared it with the baseline, percentile score. We find that our method can achieve higher correlation with human evaluation than the baseline. We picked 5 generated homophone words and replaced the original homophones with our generated ones in the source as perturbations to test the robustness of the QE systems trained by \citet{qian-etal-2024-multi,qian-etal-2024-large-language}, including fine-tuned COMETKIWI, TransQuest and MTL models as well as LLMs. At the same time, we replaced the MT output with improved translations to test how robust QE systems are towards improved translations. Our results indicate that LLMs with a larger size such as DeepSeek-67B exhibited great stability and robustness to all our perturbation groups. For future work, we plan to generate homophones at a larger scale and invite more linguists to evaluate their usefulness in real-world scenarios on social media. 

\section{Limitations and Ethical Considerations} \label{limitations}

Due to the size of the HADQAET dataset, only 167 samples that contain 5 most frequent words causing translation errors were selected in the paper. This size of test set is comparatively smaller than other robustness tests. We will generate more homophone words for testing in our future work.

The experiments in the paper were conducted using publicly available datasets. New data were created based on those publicly available datasets using computer algorithms. No ethical approval was required. The use of all data in this paper follows the licenses in~\citep{qian-etal-2023-evaluation}. 


\bibliography{anthology,custom}

\appendix

\section{Appendix} \label{sec:appendix}

\begin{CJK*}{UTF8}{gbsn}
\begin{table*}[h]
\centering
\resizebox{5cm}{!}{%
\begin{tabular}{ccc}
\toprule
\multicolumn{1}{l}{Original} & Generated & Avg Score \\ \hline
\multirow{5}{*}{尼玛} & 你妈 & 5.00 \\
 & 尼妈 & 3.75 \\
 & 泥马 & 3.50 \\
 & 尼马 & 2.75 \\
 & 泥玛 & 2.50 \\ \hdashline
\multirow{5}{*}{特么} & 他妈 & 5.00 \\
 & 她妈 & 5.00 \\
 & 它妈 & 4.00 \\
 & 踏妈 & 3.50 \\
 & 他玛 & 1.50 \\ \hdashline
\multirow{5}{*}{卧槽} & 我操 & 5.00 \\
 & 我艹 & 5.00 \\
 & 窝艹 & 3.75 \\
 & 窝操 & 3.25 \\
 & 我草 & 3.25 \\ \hdashline
\multirow{5}{*}{劳资} & 老子 & 5.00 \\
 & 老资 & 3.50 \\
 & 老自 & 2.00 \\
 & 劳子 & 1.75 \\
 & 劳自 & 1.50 \\ \hdashline
\multirow{5}{*}{草泥马} & 艹泥马 & 5.00 \\
 & 操你妈 & 4.50 \\
 & 艹你妈 & 4.50 \\
 & 草你妈 & 3.75 \\
 & 草尼妈 & 3.75 \\
 \bottomrule
\end{tabular}%
}
\caption{Original vs our generated top 5 homophone words and their average human evaluation scores (Avg Score) with and without context.}
\label{tab:generated-homo}
\end{table*}
\end{CJK*}

\begin{table*}[h]
\centering
\resizebox{15cm}{!}{%
\begin{tabular}{ccccccccccc}
\toprule
\multirow{2}{*}{Groups} & \multicolumn{2}{c}{FT-COMETKIWI} & \multicolumn{2}{c}{FT-TransQuest} & \multicolumn{2}{c}{CFT-TransQuest} & \multicolumn{2}{c}{MTL-XLM-V$_{base}$} & \multicolumn{2}{c}{MTL-XLM-R$_{large}$} \\
 & $\rho$ & $r$ & $\rho$ & $r$ & $\rho$ & $r$ & $\rho$ & $r$ & $\rho$ & $r$ \\ \hline
G0 & 0.2617 & 0.3211 & 0.2518 & 0.2954 & 0.2853 & 0.3219 & 0.2179 & 0.2139 & 0.1958 & 0.1841 \\
M1G1 & 0.2523 & 0.3014 & 0.2664 & 0.3028 & 0.2609 & 0.2987 & 0.2139 & 0.2247 & 0.1913 & 0.0076 \\
M1G2 & 0.2604 & 0.3081 & 0.2725 & 0.3144 & 0.2709 & 0.3061 & 0.2475 & 0.2358 & 0.1904 & 0.1419 \\
M1G3 & 0.2694 & 0.3276 & 0.2537 & 0.3019 & 0.2583 & 0.3013 & 0.2207 & 0.2070 & 0.1848 & 0.1865 \\
M1G4 & 0.2770 & 0.3315 & 0.2698 & 0.3238 & 0.2380 & 0.2788 & 0.2439 & 0.2284 & 0.1915 & 0.1861 \\
M1G5 & 0.2652 & 0.3100 & 0.2770 & 0.3183 & 0.2451 & 0.2806 & 0.2181 & 0.2239 & 0.1974 & 0.0935 \\
\bottomrule
\end{tabular}%
}
\caption{Original Spearman $\rho$ and Pearson's $r$ correlation scores of the perturbation groups in Method 1 on fine-tuned COMETKIWI (FT-COMETKIWI), TransQuest (FT-TransQuest) and continued fine-tuned TransQuest (CFT-TransQuest) models and multi-task learning (MTL) models based on XLM-V$_{base}$ and XLM-R$_{large}$.}
\label{tab:perturb1_ft_orig}
\end{table*}

\begin{table*}[h]
\centering
\resizebox{13cm}{!}{%
\begin{tabular}{ccccccccc}
\toprule
\multirow{2}{*}{Groups} & \multicolumn{2}{c}{Mixtral 8x7B} & \multicolumn{2}{c}{Deepseek-67B} & \multicolumn{2}{c}{FT-Yi-34B} & \multicolumn{2}{c}{FT-Deepseek-67B} \\
 & $\rho$ & $r$ & $\rho$ & $r$ & $\rho$ & $r$ & $\rho$ & $r$ \\ \hline
G0 & 0.1886 & 0.1984 & 0.2073 & 0.1338 & 0.3413 & 0.3485 & 0.2802 & 0.2469 \\
M1G1 & 0.0910 & -0.0625 & 0.1899 & 0.0878 & 0.4133 & 0.4223 & 0.2320 & 0.2756 \\
M1G2 & 0.0755 & -0.0625 & 0.1533 & 0.0609 & 0.2620 & 0.2689 & 0.2678 & 0.2444 \\
M1G3 & 0.0538 & 0.0817 & 0.1977 & 0.1441 & 0.2919 & 0.3049 & 0.2619 & 0.2780 \\
M1G4 & 0.0751 & 0.0310 & 0.0905 & 0.0047 & 0.2838 & 0.2825 & 0.3528 & 0.3833 \\
M1G5 & 0.1218 & -0.0624 & 0.1289 & 0.1342 & 0.1880 & 0.2216 & 0.2962 & 0.3305 \\
\bottomrule
\end{tabular}%
}
\caption{Original Spearman $\rho$ and Pearson's $r$ correlation scores of the perturbation groups in Method 1 on \textbf{LLMs} and \textbf{fine-tuned (FT) LLMs} as listed in Section~\ref{QE_models}.}
\label{tab:perturb1_llms_orig}
\end{table*}

\end{document}